\documentclass[lettersize,journal,table,xcdraw]{IEEEtran}
\usepackage{amsmath,amsfonts}
\usepackage{algorithmic}
\usepackage{algorithm}
\usepackage{array}
\usepackage[caption=false,font=normalsize,labelfont=sf,textfont=sf]{subfig}
\usepackage[colorlinks=true, citecolor=green, linkcolor=red]{hyperref}
\usepackage{textcomp}
\usepackage{stfloats}
\usepackage{url}
\usepackage{verbatim}
\usepackage{graphicx}
\usepackage{cite}
\usepackage{times}
\usepackage{epsfig}
\usepackage{amssymb}
\usepackage{pifont}
\usepackage{eso-pic}
\usepackage{multirow} 
\usepackage{booktabs}
\usepackage[switch]{lineno}
\usepackage{float}
\usepackage{enumitem}
\newcommand{\ie}{\emph{i.e.}}
\newcommand{\eg}{\emph{e.g.}}

\begin{document}

\title{EmotionGesture: Audio-Driven Diverse Emotional Co-Speech 3D Gesture Generation}

\author{Xingqun Qi*, Chen Liu*, Lincheng Li, Jie Hou, Haoran Xin, Xin Yu$^\dagger$
\thanks{ * these authors contributed equally to this work. This work was done when Xingqun Qi was an intern at the NetEase Fuxi AI Lab, Hangzhou, China.}
\thanks{$^\dagger$ corresponding author: xin.yu@uq.edu.au}
\thanks{Xingqun Qi is with the Academy of Interdisciplinary Studies, The Hong Kong University of Science and Technology, Hong Kong, China. (e-mail: xingqun.qi@connect.ust.hk).}
\thanks{Chen Liu and Xin Yu are with the School of Electrical Engineering and Computer Science, The University of Queensland, Queensland, Australia. (e-mail: uqcliu32@uq.edu.au, xin.yu@uq.edu.au)}
\thanks{Lincheng Li, Jie Hou, and Haoran Xin are with the NetEase Fuxi AI Lab, Hangzhou, China. (e-mail: lilincheng@corp.netease.com, houjie1@corp.netease.com, xinhaoran@corp.netease.com)
}

}

\markboth{Journal of \LaTeX\ Class Files,~Vol.~14, No.~8, August~2021}%
{Shell \MakeLowercase{\textit{et al.}}: A Sample Article Using IEEEtran.cls for IEEE Journals}


\maketitle

\begin{abstract}
Generating vivid and diverse 3D co-speech gestures is crucial for various applications in animating virtual avatars. While most existing methods can generate gestures from audio directly, they usually overlook that \textbf{\emph{emotion}} is one of the key factors of authentic co-speech gesture generation. In this work, we propose \textbf{\emph{EmotionGesture}}, a novel framework for synthesizing vivid and diverse emotional co-speech 3D gestures from audio. Considering emotion is often entangled with the rhythmic beat in speech audio, we first develop an Emotion-Beat Mining module (EBM) to extract the emotion and audio beat features as well as model their correlation via a transcript-based visual-rhythm alignment. Then, we propose an initial pose based Spatial-Temporal Prompter (STP) to generate future gestures from the given initial poses. STP effectively models the spatial-temporal correlations between the initial poses and the future gestures, thus producing the spatial-temporal coherent pose prompt. Once we obtain pose prompts, emotion, and audio beat features, we will generate 3D co-speech gestures through a transformer architecture. However, considering the poses of existing datasets often contain jittering effects, this would lead to generating unstable gestures. To address this issue, we propose an effective objective function, dubbed Motion-Smooth Loss. Specifically, we model motion offset to compensate for jittering ground-truth by forcing gestures to be smooth. Last, we present an emotion-conditioned VAE to sample emotion features, enabling us to generate diverse emotional results. Extensive experiments demonstrate that our framework outperforms the state-of-the-art, achieving vivid and diverse emotional co-speech 3D gestures. Our code and dataset will be released at the project page: \href{https://xingqunqi-lab.github.io/Emotion-Gesture-Web/}{\textit{https://xingqunqi-lab.github.io/Emotion-Gesture-Web/}}.
\end{abstract}


\section{Introduction}
\IEEEPARstart{C}{o-speech} 3D gesture generation aims to synthesize vivid and diverse human poses consistent with the corresponding audio. This non-verbal body language helps people express their views and ideas more comprehensibly in daily communication~\cite{cassell1999speech,wagner2014gesture,de2012interplay}. Thus, animating virtual avatars with audio-coherent human gestures is crucial for various applications in embodied AI agents~\cite{huang2012robot, Gadre_2022_CVPR, padmakumar2022teach, wolfert2022review, jahanmahin2022human}. Conventionally, recent researchers build the end-to-end mapping between the speech audio and corresponding upper body dynamics~\cite{liang2022seeg, liu2022learning, zhu2023taming, yi2022generating, liu2022disco}. They usually leverage a few initial postures as reference prompts to guide the generation~\cite{Ao2023GestureDiffuCLIP}.

Most existing co-speech gesture generation methods address this challenging task by constructing a large corpus and then modeling the correlation between audio and gestures~\cite{yi2022generating, liu2022disco, liu2022audio, ao2022rhythmic}. A few pioneering researchers have modeled emotional co-speech gestures by simply adding the emotion as a one-hot condition~\cite{liu2022beat, yang2023DiffuseStyleGesture}. These works overlook effectively exploring the emotional information of audio signals, resulting in unrealistic gestures in most real-world scenarios. Moreover, since the pose annotations of existing datasets~\cite{yoon2019robots,yoon2020speech,liu2022learning} are often obtained by pre-trained 3D pose estimators, their poses usually contain jittering effects. 
Directly regressing postures from ground-truth might not yield smooth synthetic gesture sequences, as in previous studies~\cite{liang2022seeg, liu2022learning, zhu2023taming}.

As evidenced in previous works, there are three main challenges in this task: 
\begin{itemize}
\item \emph{How to effectively model the diverse emotional co-speech gestures?}

\item \emph{How to enforce the co-speech gestures to be well aligned with audio beats?}

\item \emph{How to achieve posture spatial-temporal smoothness when ground-truth 3D gestures are jittery?}
\end{itemize}

\begin{figure*}[t]
\begin{center}
\includegraphics[width=1\linewidth]{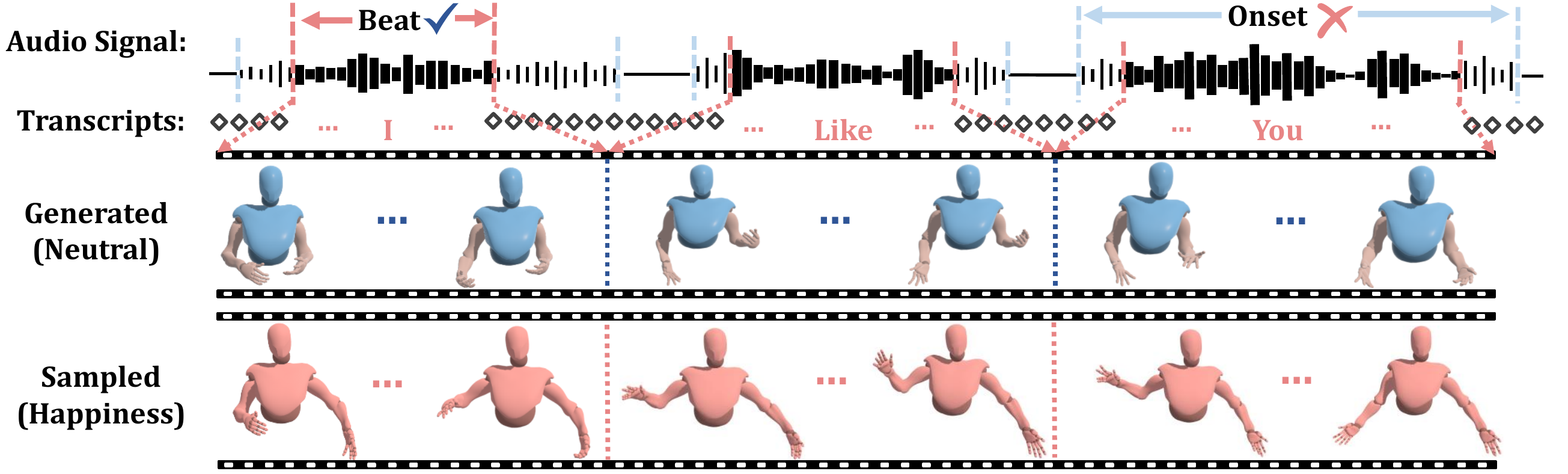}
\end{center}
\caption{Diverse \textbf{emotional} exemplary clips sampled by our \textbf{\emph{EmotionGesture}}, the pause duration in transcript is padded as $\left ( \diamond  \right )$. We identify the \textbf{beat} via frame-wise aligned \textbf{utter words} (\textcolor[RGB]{232, 132, 132}{pink}) in audio-synchronized transcripts. Due to the noisy environment, it is improper to directly extract \textbf{audio onsets} (\textcolor{blue}{blue vertical lines in audio signal}) as rhythmic indicators.}
\label{fig:figure1}
\end{figure*}

In this work, we propose a novel framework, \textbf{\emph{EmotionGesture}}, to generate vivid and diverse emotional 3D co-speech gestures driven by audio. In our EmotionGesture, we first propose an Emotion-Beat Mining (EBM) module to model emotional co-speech gestures. 
EBM extracts the emotion and audio beat features from the input audio signals. 
To achieve the rhythmic beats, previous works~\cite{liang2022seeg, ao2022rhythmic} leverage the audio onset as an indicator. However, onset extraction may be affected by audio noise, thus producing low-fidelity audio-driven gestures. We observe that utter words with frame-wise timestamps can be employed to align the beats for the extracted audio beat features from BEM, as depicted in Figure~\ref{fig:figure1}. 
Hence, we design a contrastive learning fashion to enforce the beat features to be frame-wise aligned with audio rhythm through the synchronized transcripts.
For emotion feature extraction, we employ an emotion classifier to ensure the extracted feature can represent the emotion in the audio.

Then, we propose an initial pose based Spatial-Temporal Prompter (STP) to generate future poses upon the initial poses. STP aims to ensure smoothness between the initial postures and future poses via prompt enhancement. 
Due to the mismatched temporal dimension of initial poses and target ones, previous works~\cite{yoon2020speech, liu2022learning, zhu2023taming} straightly pad the temporal dimension of initial postures as reference prompt, which might result in unnatural and ambiguous gestures.
Instead of this, our STP fully takes advantage of the motion clues from the initial poses to produce future postures via two prompt learning strategies: spatial-interpolation prompt learning and temporal-reinforcement prompt learning. 
The spatial-interpolation prompt learning strategy leverages the embedding of initial postures as guidance to update each future frame via a learnable interpolation manner. 
It provides spatial-wise smoothness but fails in the long-term sequential constraints.
As a complement, the temporal-reinforcement prompt learning strategy aggregates historical temporal changes to consolidate the temporal correlation for future sequential steps. Afterward, we concatenate the initial pose features and future pose features as the enhanced pose prompt to guide the smooth gesture generation.

Furthermore, considering the jittering issue of pose annotations,
we design a simple yet effective objective function, dubbed Motion-Smooth Loss, inspired by soft label smoothing technique \cite{hinton2015distilling, allen2020towards}. Here, we produce the smoothed motion offset of the ground truth through a temperature coefficient. Then, the smoothed motion offset is applied to provide supervision of generating temporally smoothed gestures. 
In addition, we introduce an emotion-conditioned VAE~\cite{sohn2015learning} to sample the diverse emotion features based on these emotion-clustered features. In this fashion, our framework enables diverse emotional gesture generation, as displayed in Figure~\ref{fig:figure1}.

Moreover, the current largest emotional co-speech gesture dataset~\cite{liu2022beat} only includes 30 avatar identities with limited pre-defined talking topics. This may lead to insufficient diversity of 3D postures and speech content. Therefore, we newly collected a TED Emotion dataset composed of more than 1.7K avatar identities from TED talk show videos. 
Extensive experiments on BEAT and TED Emotion datasets demonstrated that our \textbf{\emph{EmotionGesture}} significantly outperforms various counterparts, displaying vivid and diverse emotional co-speech 3D gestures. 

To summarize, our main contributions are four-fold: 
\begin{itemize}
\item We devise EmotionGesture, a novel framework that achieves audio-driven diverse emotional co-speech 3D gesture generation.

\item We propose an Emotion-Beat Mining (EBM) module to facilitate diverse emotional gesture generation while aligning generated gestures with audio beats.

\item We present a Spatial-Temporal Prompter (STP) to obtain the enhanced temporal-coherency pose prompt, thus guiding the smooth gesture generation.

\item We design a simple yet effective Motion-Smooth Loss to overcome the pose jittering issues in existing datasets, thus achieving temporally smooth co-speech gestures.
\end{itemize}

\section{Related Work}

\subsection{Co-speech Gesture Generation.}
Co-speech gesture generation aims at synthesizing audio-synced human pose sequences of the talking people. It has witnessed impressive research interests for its wide practical value in various applications like human-agent interaction~\cite{koppula2013anticipating, huang2022proxemics, salem2012generation, wang2022self}, robotics~\cite{ishi2018speech, yoon2019robots}, and holoportation~\cite{orts2016holoportation}. Thus, numerous studies have been proposed to tackle this challenging task. Traditionally, previous researchers follow the rule-based pipelines that the mapping between speech and gestures is pre-defined by linguistic experts~\cite{cassell1994animated, huang2012robot, marsella2013virtual, poggi2005greta}. In this pattern, researchers focus on refining the transitions matching process between the generated different motions. However, it may need expensive efforts for the experts when facing complex scenes.

Recently, thanks to more and more released datasets~\cite{yoon2019robots, yoon2020speech, liu2022learning, liu2022beat}, co-speech gesture generation is significantly improved by deep-learning based approaches. Among various approaches, many researchers intend to utilize multi-modality clues to build the associations between co-speech gestures and audio signals~\cite{li2021audio2gestures}, text transcripts~\cite{ahuja2019language2pose}, and speaker identity~\cite{yoon2020speech, liu2022learning}. Only a few research~\cite{liu2022beat, Ao2023GestureDiffuCLIP} simply explore the significance between generated gestures and emotions. However, they fail to fully exploit the emotional pheromones of speech audio, resulting in lower emotion-related and unrealistic gestures in most real-world scenarios. In this work, we propose to extract the diverse emotion representation from input speech audio, thus achieving audio-driven emotion control of the generated gestures.

\subsection{3D Human Motion Modeling.}
Modeling 3D human motion plays a critical role in research areas of both computer vision and graphics. One of the key targets is to preserve the spatial-temporal coherency of the generated human motions~\cite{battan2021glocalnet, zhong2022spatio, mao2022contact, mao2022weakly}. To achieve that, previous studies~\cite{akhter2008nonrigid, huang2017towards, mao2019learning, zhao2023poseformerv2} directly employ the discrete cosine transform (DCT) as the post-processing strategy. However, these straightforward approaches fail in the jittering problem in pseudo-annotated 3D co-speech gesture datasets. 
Motivated by these methods, we propose a spatial-temporal prompter to keep the smoothness between the initial poses and generated gestures. In addition, we design a simple yet effective motion-smooth loss function to smooth our generated co-speech motions. Notably, our motion-smooth loss gives a practical solution to address the jittering issue. Such design could prospectively provide insights into relevant domains not only on co-speech gesture generation but also on 3D human pose estimation~\cite{zheng20213d}, motion prediction\cite{tevet2022human}, and talking head synthesis\cite{huang2022audio}.

\section{Proposed Method}

\subsection{Problem Formulation}

Given speech audio sequence $A = \left \{ a_{1},...,a_{N} \right \}$ as input, the goal of our framework $\mathcal{G}$ is to generate continuous diverse emotional 3D co-speech poses as $P = \left \{ p_{1},...,p_{N} \right \}$, where $N$ denotes the total frame number corresponding to $A$. Here, $p_{i}$ represents $J$ joints of the human upper body including two hands. To ensure that the generated co-speech gestures are diverse emotional while preserving alignment with the audio beat, we also introduce the emotion label and text transcripts $T = \left \{ t_{1},...,t_{N} \right \}$ to provide supervision in the training phase. Notice that the emotion label actually is the one-hot vector in our framework. With the aforementioned representations, the overall objective is expressed as:
\begin{equation}
\underset{\mathcal{G}}{argmin} \left \| P-\mathcal{G} \left ( A, \left \{ p_{1},...,p_{M}\right \}  \right ) \right \|,
\label{Eq1}
\end{equation}
where $\left \{ p_{1},...,p_{M} \right \}$ is initial pose sequence.

\begin{figure*}[t]
\begin{center}
\includegraphics[width=1\linewidth]{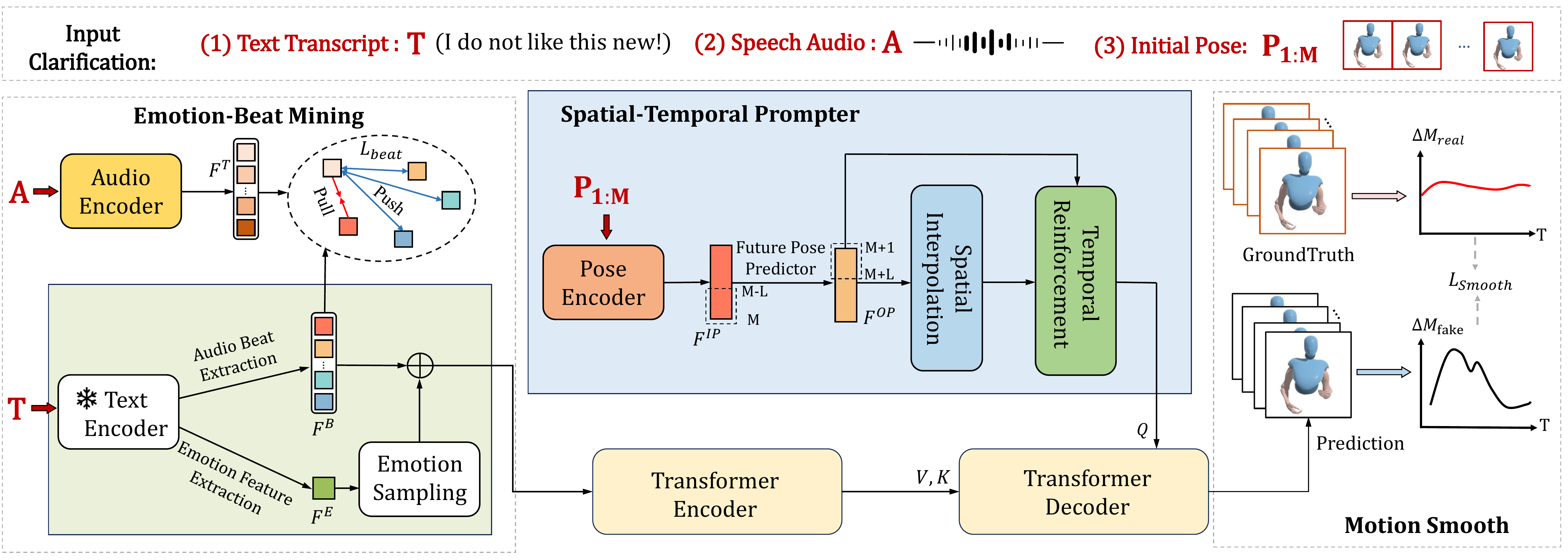}
\end{center}
\caption{Overview of our proposed \textbf{\emph{EmotionGesture}} framework. With extracted audio beat features $F^{B}$, and emotion features $F^{E}$, we could achieve the generation of audio-driven diverse emotional co-speech gestures. Our spatial-temporal prompter aims to obtain the enhanced temporal-coherency pose prompt based on the initial pose sequence. }
\label{fig:figure2}
\end{figure*}

\subsection{Emotion-Beat Mining}
To achieve the audio-driven emotion control on the generated diverse co-speech gesture and maintain the rhythmic alignment with speech audio, the inherently entangled emotion features and rhythmic beat have to be extracted independently from the input speech audio. Thus, we propose an Emotion-Beat Mining (EBM) module to extract the features of emotion and audio beat. As illustrated in Figure~\ref{fig:figure2}, we leverage an audio encoder $E_{a}$ combined with two MLP-based projectors to obtain two separate features, \ie, audio beat features $F^{B}=\left \{ f_{1}^{B},...,f_{N}^{B}  \right \}$ and emotion features $F^{E}$.

\subsubsection{Beat Alignment}
We intend to enforce the embedded beat features being frame-wise aligned with the speech rhythm. Intuitively, previous works utilize the signal processing technique~\cite{bello2005tutorial} directly identify the onsets\footnote{Onset refers to the beginning or starting point of a sound signal~\cite{bello2005tutorial}.} of audio signals as the speech rhythm~\cite{ao2022rhythmic,liang2022seeg}.
Instead of this, our beat-alignment strategy is built upon the insight that ``\textbf{\emph{beat starts when people are speaking}}". Thus, we introduce the text transcripts synchronized with frame-wise timestamps (\ie, each uttered word is aligned with a frame-wise gesture) to provide beat alignment supervision via contrastive learning. 

In the audio-synchronized test transcripts, the pause duration is inserted with padding tokens 
while the uttered words are encoded as unique identifiers. As shown in Figure~\ref{fig:figure2}, we leverage a pre-trained word2vec model~\cite{mikolov2013efficient} as our text encoder $E_{t}$ to obtain the audio-coherency transcript features $F^{T}=\left \{ f_{1}^{T},...,f_{N}^{T}  \right \}$. In our contrastive learning formulation, we utilize the uttered word feature $f_{u}^{T}$ as the anchor sample, and only the beat feature aligned with the current uttered word serves as a positive sample, denoted as $f_{u}^{B+}$. Then, the negative samples are defined as other timesteps beat features in the same sequence. In this manner, beat features of other timesteps that reflect different rhythmic and semantic information are repelled. Drawing inspiration from InfoNCE~\cite{oord2018representation}, our beat-alignment contrastive learning loss is expressed as:
\begin{align}
\mathcal{L}_{beat}& =  
\notag
\\& \mathbb{I}_{\left [ U\ne 0 \right ] } \left ( -log {\textstyle \sum_{u=1}^{U}} \frac{  exp\left ( \mathit{sim}\left ( f_{u}^{T}, f_{u}^{B+} \right ) / \tau   \right ) }{ {\textstyle \sum_{i=1, i\ne u}^{N-1}}exp\left ( \mathit{sim}\left ( f_{u}^{T}, f_{i}^{B-} \right ) / \tau  \right )  } \right ) ,
\label{Eq2}
\end{align}
where $U$ is the total number of  uttered words, $\mathbb{I}_{\left [ U\ne 0 \right ] }\in \left [ 0,1 \right ]$ is an indicator function evaluating to 0 if there are no utterances in this audio duration. $\mathit{sim}\left (\cdot \right )$ denotes cosine similarity, and $\tau$ is the temperature hyperparameter.
Different from the HA2G model~\cite{liu2022learning} directly taking the sequence-wise words features $F^{T}$ to constrain the audio embedding features $F^{B}$ keeping semantic consistency, globally. Our contrastive learning loss aims to achieve frame-wise beat alignment. Meanwhile, we effectively prevent model collapse when there are no uttered words in the audio duration (\ie, no rhythmic onsets). 

\subsubsection{Emotion Sampling} 
\begin{figure*}[t]
\begin{center}
\includegraphics[width=0.95\linewidth]{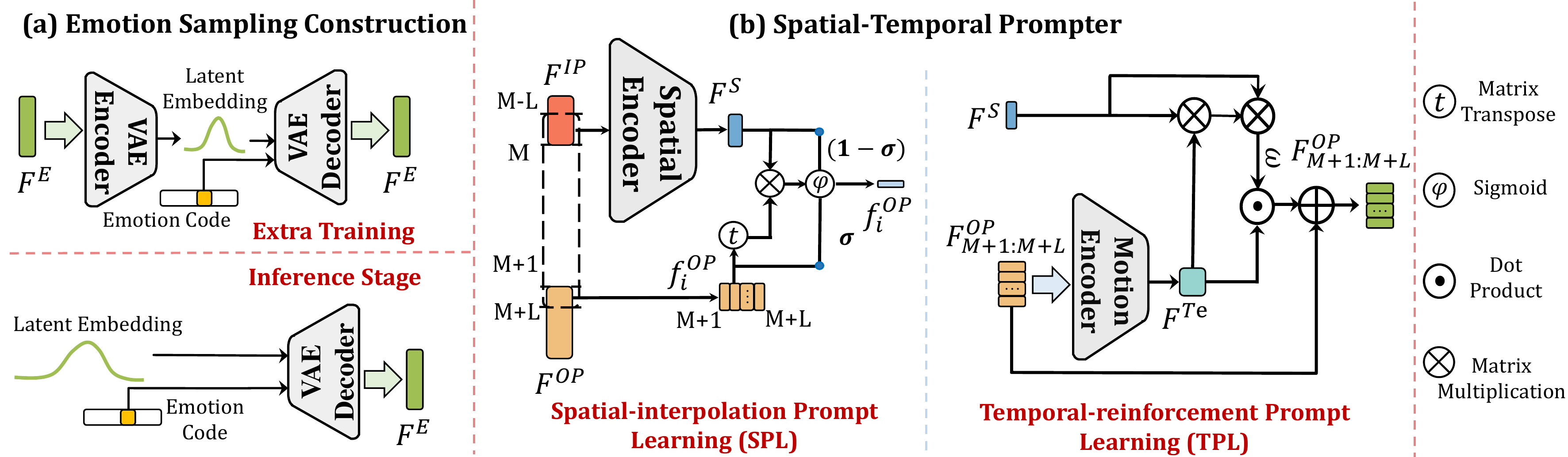}
\end{center}
\caption{Details of our proposed emotion sampling component and spatial-temporal prompter. (a) In the emotion sampling process, we introduce an emotion-conditioned VAE to obtain the diverse emotion embeddings $F^{E}$. Emotion code is the one-hot vector representing the different emotions. (b) The Spatial-Temporal Prompter aims to provide the temporal-coherency pose prompt to guide gesture generation. }
\label{fig:figure3}
\end{figure*}

In order to fully extract the emotional features, we leverage the emotion label to provide the classification supervision via a classification header. The supervision is formulated as the cross-entropy loss: $\mathcal{L}_{emo}=-\sum_{c=1}^{C}  y_{c}\cdot log \left (q_{c}   \right ) $, where $C$ represents the number of emotions, $y_{c}$ denotes whether the sample belongs to emotion label $c$, and $q_{c}$ is the prediction probability. 
Afterward, to achieve the diverse emotion gesture generation, once our EmotionGesture is well trained, we freeze the parameters of the overall framework and train an emotion-conditioned VAE~\cite{sohn2015learning} model upon clustered-emotion features $F^{E}$ as depicted in Figure \ref{fig:figure3} (a: Extra Training). Once we obtain sampled diverse emotion features, we feed the summation of emotion features and beat features into the transformer-based backbone to produce co-speech 3D gestures. In the inference phase, the emotion-conditioned VAE samples diverse emotion features to produce diverse emotional gestures, as depicted in Figure~\ref{fig:figure3} (a).

\subsection{Spatial-Temporal Prompter}
Conventionally, researchers directly take the padded initial poses as the conditioned prompt to guide the co-speech gesture synthesis.  
Unlike the previous works~\cite{yoon2020speech,zhu2023taming,liu2022learning}, we aim to build the enhanced temporal-coherency pose prompt based on the initial pose sequence. This pose prompt would lead to smooth co-speech gesture generation. Thus, we propose a Spatial-Temporal Prompter (STP) for keeping smoothness among the enhanced pose prompt.
As shown in Figure~\ref{fig:figure2}, we first utilize a pose encoder to obtain the initial pose embedding $F^{IP}=\left \{ f_{1}^{IP},...,f_{M}^{IP}  \right \}$.
Then, we produce the future pose features $F^{OP}=\left \{ f_{M+1}^{OP},...,f_{N}^{OP}  \right \}$ via a 1-D convolution-based pose predictor. Inspired by \cite{mao2022weakly}, we nominate a transition chunk consisting of the last $L$ frames of initial pose embeddings and the first $L$ ones of the future pose features. Here, the dimension of each frame in the transition chunk is $\mathbb{R}^{1\times D}$.
STP keeps the transition from the initialized $L$ poses to the future $L$ poses to be a smooth sequence. Concretely, STP consists of two prompt learning strategies, \ie, spatial-interpolation prompt learning and temporal-reinforcement prompt learning.

\subsubsection{Spatial-interpolation Prompt Learning} 
As shown in Figure~\ref{fig:figure3} (b), our spatial-interpolation prompt learning strategy aims to ensure spatial-wise smoothness between the initialized $L$ frames and the future $L$ frames in a learnable interpolation pattern. Concretely, we first design a spatial encoder to obtain the ensembled spatial representation of the initialized $L$ poses, denoted as $F^{S}\in \mathbb{R}^{1\times D} $. Then, we calculate the spatial interpolation score between the ensembled spatial representation and each frame in the future $L$ poses. With the help of the interpolation score, each spatial smoothed feature of the future $L$ frames is represented as: 
\begin{align}
f_{i}^{OP} & = \sigma f_{i}^{OP} + (1-\sigma)F^{S}, 
\notag
\\ \sigma  &= \varphi \left ( F^{S} \otimes (f_{i}^{OP})^{'} \right ) ,
\notag
\\ i & \in \left [ M+1,M+L \right ] ,
\label{Eq3}
\end{align}
where $\sigma$ is the interpolation score, $\varphi$ is the sigmoid operation, $\otimes$ is matrix multiplication, $'$ indicates the transpose operation.

\subsubsection{Temporal-reinforcement Prompt Learning} 
To further enhance the long-term spatial-temporal smoothness in the transition chunk, we propose a temporal-reinforcement prompt learning strategy. Specifically, we develop a motion encoder to obtain the sequence-aware temporal embedding of the future $L$ pose features, denoted as $F^{Te}\in \mathbb{R}^{L\times 1} $. Here, the temporal embedding represents the temporal changes among predicted $L$ poses. Then, we compute the long-term spatial-temporal correlation score between the ensembled spatial representation and temporal embedding, globally. Once we obtain this long-term correlation score, the temporal-reinforced future $L$ poses sequence in transition is attained by:
\begin{align}
F_{M+1:M+L}^{OP} &=  F_{M+1:M+L}^{OP} + F_{M+1:M+L}^{OP}\odot \omega, 
\notag
\\ \omega &= \mathcal{S} \left (  (F^{Te}\otimes F^{S})\otimes F^{S} \right ),
\label{Eq4}
\end{align}
where $\odot$ denotes dot product, $\mathcal{S}$ means softmax operation, $\omega\in \mathbb{R} ^{L\times 1}$ is the calculated long-term spatial-temporal correlation score. Afterward, we concatenate the initial pose embedding $F^{IP}$ and future pose features $F^{OP}$ as the enhanced pose prompt feed to the decoder. 

To fully take advantage of the initial poses, we leverage the enhanced pose prompt as the query $Q$. Then, we use $Q$ to match the key features $K$ and value features $V$ in the transformer-based decoder via three times Multi-Head Attention (MHA) \cite{vaswani2017attention}, expressed as:
\begin{align}
MultiHead(Q, K, V) = softmax(\frac{QK }{\sqrt{d} } )V.
\end{align}
In this fashion, sequence-aware correspondence between the emotional audio representation and pose prompt is jointly built. Then, similar to~\cite{ng2021body2hands}, we employ a motion discriminator to ensure the generated co-speech gestures preserve realism.

\begin{table*}[t]
\centering
\caption{The accuracy of our pre-trained emotion classifier on the BEAT dataset.}
\label{table4}
\renewcommand{\arraystretch}{1.3} 
\begin{tabular}{cccccccccc}
\toprule
Emotion & Neutral & Anger & Happiness & Fear & Disgust & Sadness & Contempt & Suprise & Average \\ \midrule
Accuracy(\%) & 99.91 & 98.80 & 99.84 & 99.60 & 100.00 & 99.67 & 99.75 & 98.30 & \textbf{99.70} \\ \bottomrule
\end{tabular}%
\end{table*}

\subsection{Training Objectives}

\subsubsection{Motion-Smooth Loss.} 
To address the jittering problem in most existing co-speech 3D gesture datasets, we design a simple yet effective objective, named Motion-Smooth Loss. Our motion-smooth loss aims to produce the smoothed motion offset as the target. Concretely, we first compute the motion offset of the jittering ground truth as $\bigtriangleup \mathcal{M}_{real}\in  \mathbb{R}^{(N-1)\times D}$. Meanwhile, we obtain the motion offset of the generated gestures as $\bigtriangleup \mathcal{M}_{fake}\in  \mathbb{R}^{(N-1)\times D}$ in the same way. Inspired by soft label smoothing technique~\cite{hinton2015distilling, allen2020towards}, we leverage a smooth temperature coefficient to produce the smoothed ground truth. Then, our motion-smooth loss is formulated as:
\begin{align}
\mathcal{L} _{smooth} & = \mathcal{KL} \left ( \mathcal{S} \left ( \bigtriangleup \mathcal{M}_{real}/\Gamma  \right ) \left |  \right | \mathcal{S} \left ( \bigtriangleup \mathcal{M}_{fake} \right ) \right ), 
\label{Eq5}
\end{align}
where $\Gamma$ is the smooth temperature coefficient, $\mathcal{S}$ means softmax operation, $\mathcal{KL}$ denotes Kullback Leibler Divergence.

\subsubsection{Reconstruction Loss.} 
We leverage the ground truth gestures to constrain the generated co-speech gestures as:
\begin{align}
\mathcal{L}_{rec} \! & = \! \left \| P\!-\! \tilde{P}\right \| _{1},
\label{Eq6}
\end{align}
where $\tilde{P}$ denotes generated gestures.

\subsubsection{Adversarial Learning.}
Following the configuration of previous works, we employ the adversarial training loss as:
\begin{align}
\mathcal{L}_{adv} = \mathbb{E} _{P} \left [ log \mathcal{D}(P) \right ] + \mathbb{E} _{A} \left [ log(1- (\mathcal{G}\left ( A , \left \{ p_{1},...,p_{M}\right \}  \right )) \right ],
\end{align}
where $\mathcal{D}$ denotes the discriminator and $\mathcal{G}$ means generator. Finally, the \textbf{overall objective} is:
\begin{align}
\!\!\min_{\mathcal{G}}\max_{\mathcal{D}}\mathcal{L}_{total} &=  \lambda_{r}\mathcal{L} _{rec} +  \mathcal{L} _{adv} + \lambda_{b}\mathcal{L} _{beat} 
\notag
\\& + \lambda_{e} \mathcal{L} _{emo} + \lambda_{s}\mathcal{L} _{smooth},
\label{eq11}
\end{align}
where $\mathcal{L} _{beat}$ denotes our beat-alignment contrastive learning loss and $\mathcal{L} _{emo}$ indicates our emotion classification loss. The $\lambda_{r}$, $\lambda_{b}$, $\lambda_{e}$, and $\lambda_{s}$ are weight coefficients.

\begin{table*}[t]
\centering
\caption{Comparison with the state-of-the-art methods on the BEAT dataset and TED Emotion dataset of the generated co-speech gestures. $\downarrow$ indicates the lower the better, and $\uparrow$ indicates the higher the better. $\pm$ means 95\% confidence interval. $\dagger$ means our EmotionGesture framework is implied without the sampling phase during inference.}
\label{table1}
\resizebox{\textwidth}{!}{%
\renewcommand{\arraystretch}{1.3} 
\begin{tabular}{llcccccccccccc}
\toprule
 &  & \multicolumn{6}{c}{BEAT Dataset} & \multicolumn{6}{c}{TED Emotion Dataset} \\ \cmidrule(r){3-8} \cmidrule(r){9-14}
\multirow{-2}{*}{Settings} & \multirow{-2}{*}{Models} & L2$\downarrow$ & MPJRE$\downarrow$ & FGD$\downarrow$ & BA$\uparrow$ & EA(\%)$\uparrow$ & Diversity$\uparrow$ & L2$\downarrow$ & MPJRE$\downarrow$ & FGD$\downarrow$ & BA$\uparrow$ & EA(\%)$\uparrow$ & Diversity$\uparrow$ \\ \midrule
 & Seq2Seq\cite{yoon2019robots} & 2.48 & 5.90 & 1.11 & 0.30 & 65.24 & - & 2.10 & 5.70 & 1.63 & 0.50 & 66.43 & - \\
 & S2G\cite{ginosar2019learning}& 2.14 & 3.82 & 1.09 & 0.79 & 65.39 & - & 1.68 & 4.70 & 0.65 & 0.79 & 68.10 & - \\
 & JointEM\cite{ahuja2019language2pose} & 2.33 & 4.08 & 2.05 & 0.28 & 53.01 & - & 1.24 & 3.38 & 0.73 & 0.26 & 58.45 & - \\
 & CAMN\cite{liu2022beat} & 1.97 & 3.56 & 1.12 & 0.74 & 71.96 & - & 1.24 & 3.27 & 0.87 & 0.87 & 71.86 & - \\
\multirow{-5}{*}{\begin{tabular}[c]{@{}c@{}}Constant\\ Gestures\end{tabular}} & \cellcolor[HTML]{ECF4FF}\textbf{Ours $\dagger$} & \cellcolor[HTML]{ECF4FF}\textbf{1.59} & \cellcolor[HTML]{ECF4FF}\textbf{2.69} & \cellcolor[HTML]{ECF4FF}\textbf{0.47} & \cellcolor[HTML]{ECF4FF}\textbf{0.93} & \cellcolor[HTML]{ECF4FF}\textbf{81.21} & \cellcolor[HTML]{ECF4FF}\textbf{-} & \cellcolor[HTML]{ECF4FF}\textbf{0.93} & \cellcolor[HTML]{ECF4FF}\textbf{2.37} & \cellcolor[HTML]{ECF4FF}\textbf{0.11} & \cellcolor[HTML]{ECF4FF}\textbf{0.93} & \cellcolor[HTML]{ECF4FF}\textbf{85.59} & \cellcolor[HTML]{ECF4FF}\textbf{-} \\ \midrule
 & Trimodal\cite{yoon2020speech} & - & - & 2.94 & 0.86 & 32.31 & $30.52^{\pm 0.41}$ & - & - & 0.76 & 0.77 & 69.35 & $16.68^{\pm 0.48}$ \\
 & HA2G\cite{liu2022learning} & - & - & 2.45 & 0.76 & 57.10 & $11.90^{\pm0.47}$ & - & - & 0.71 & 0.75 & 72.30 & $9.10^{\pm 0.32}$ \\
 & DiffGes\cite{zhu2023taming} & - & - & 2.03 & 0.82 & 15.62 & $36.61^{\pm0.61}$ & - & - & 1.28 & 0.82 & 33.41 & $18.28^{\pm 2.42}$ \\
 & TalkShow\cite{yi2022generating} & - & - & 0.91 & 0.84 & 45.01 & $20.57^{\pm0.70}$ & - & - & 0.83 & 0.84 & 74.20 & $11.42^{\pm 0.92}$ \\
\multirow{-5}{*}{\begin{tabular}[c]{@{}c@{}}Diverse\\ Gestures\end{tabular}} & \cellcolor[HTML]{ECF4FF}\textbf{Ours} & \cellcolor[HTML]{ECF4FF}\textbf{-} & \cellcolor[HTML]{ECF4FF}\textbf{-} & \cellcolor[HTML]{ECF4FF}\textbf{0.52} & \cellcolor[HTML]{ECF4FF}\textbf{0.91} & \cellcolor[HTML]{ECF4FF}\textbf{81.16} & \cellcolor[HTML]{ECF4FF}\textbf{39.34$^{\pm 1.02}$} & \cellcolor[HTML]{ECF4FF}\textbf{-} & \cellcolor[HTML]{ECF4FF}\textbf{-} & \cellcolor[HTML]{ECF4FF}\textbf{0.13} & \cellcolor[HTML]{ECF4FF}\textbf{0.91} & \cellcolor[HTML]{ECF4FF}\textbf{84.81} & \cellcolor[HTML]{ECF4FF}\textbf{19.46$^{\pm 0.19}$} \\ \bottomrule
\end{tabular}%
}
\end{table*}

\section{Experiments}

\subsection{Datasets and Experimental Setting}

\subsubsection{BEAT Dataset.} 
BEAT~\cite{liu2022beat} is a large-scale multi-emotion dataset for conversational gestures synthesis, including audio, transcripts, and 3D human whole-body motions from 30 speakers. In BEAT, the audio and corresponding gestures are annotated with 8 emotional styles (\ie, \emph{neutral, anger, happiness, fear, disgust, sadness, contempt, and surprise}) in 4 languages. In our experiments, similar to \cite{yoon2019robots,yoon2020speech,liu2022learning}, we use the upper body with 47 joints of English speakers, amounting to about 35 hours duration. Meanwhile, we resample the human motion with 15 FPS and select the continuous 60 frames with a stride of 30 as gesture clips. Finally, we obtain 55,420 clips. The training, validation, and testing subsets are divided following the proportion as 70\%, 10\%, and 20\%. In particular, the numbers of sequences in each data partition are Training set: 38,814; Validation set: 5,502; Testing set: 11,10. Similar to \cite{liu2022beat}, we keep the partition of different emotions in each sub-dataset as:
Neutral 51\%, Anger 7\%, Happiness 7\%, Fear 7\%, Disgust 7\%, Sadness 7\%, Contempt 7\%, and Suprise 7\%.


\subsubsection{TED Emotion Dataset.} 
Inspired by \cite{liu2022learning, yi2022generating}, we newly collect a TED Emotion dataset based on in-the-wild TED talk show videos. In particular, we leverage the state-of-the-art human pose estimator ExPose~\cite{choutas2020monocular} to obtain the 43 3D upper body joints as pseudo ground truth. Then we keep a similar data processing strategy with the BEAT dataset to acquire 107,468 gesture clips, and each clip has 60 frames. To acquire the emotion label of the TED Emotion dataset, we leverage the BEAT to pre-train an audio-based emotion classifier. Concretely, we leverage the BEAT training set to pre-train the classifier, and the validation set to select the optimal model. Finally, the emotion classification accuracy on the BEAT testing set is 99.70\%. For each emotion, the accuracy is reported in Table~\ref{table4}. 

Then, we annotate the emotion label of the TED Emotion dataset via this emotion classifier. 
To ensure the annotation accuracy of emotion categories, we randomly visualize gestures for each emotion category and set the classification threshold $\ge 0.9$. Meanwhile, we drop the two uncommon emotions, \emph{disgust and contempt}. Finally, we totally gain 78,734 gesture clips. Our experiments adopt the same division criteria as \cite{liu2022learning} to split the training, validation, and testing subsets. The numbers of sequences in each data partition of our newly annotated TED Emotion dataset are Training set: 66,679; Validation set: 6,258; Testing set: 5,797. Our newly collected TED Emotion Dataset is released at \href{https://github.com/XingqunQi-lab/EmotionGestures}{\textit{https://github.com/XingqunQi-lab/EmotionGestures}}.

\begin{figure*}
\begin{center}
\includegraphics[width=0.95\linewidth]{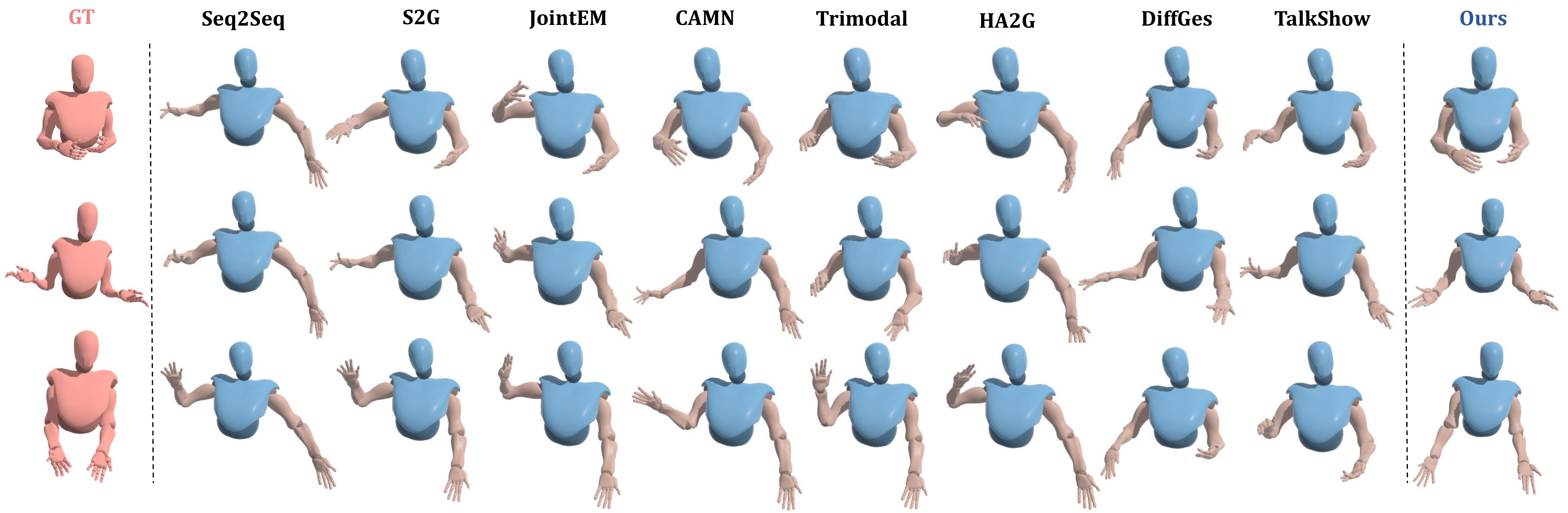}
\end{center}
\caption{ Visualization of our predicted 3D hand gestures against various state-of-the-art methods~\cite{yoon2019robots, ginosar2019learning, ahuja2019language2pose, liu2022beat, yoon2020speech, liu2022learning, zhu2023taming, yi2022generating}. From top to bottom, we show the three keyframes (an early, a middle, and a late one) of a pose sequence.
Best view on screen.}
\label{fig:figure4}
\end{figure*}

\begin{table*}[t]
\centering
\caption{Ablation study on different loss functions and different components of our proposed EmotionGesture framework. $\downarrow$ indicates the lower the better, and $\uparrow$ indicates the higher the better. $\pm$ means 95\% confidence interval. $\ddagger$ denotes we directly concatenate the one-hot emotion label with audio features as input. "Spatial" means the spatial-interpolation prompt learning, and "Temporal" denotes the temporal-reinforcement prompt learning. "+" indicates that we continue to add the corresponding component or loss function upon "Baseline", sequentially.
Notice that only with the sampling setting our framework could achieve diverse co-speech gesture synthesis. }
\label{table2}
\renewcommand{\arraystretch}{1.2} 
\begin{tabular}{lcccccccccccc}
\toprule
 & \multicolumn{6}{c}{BEAT Dataset} & \multicolumn{6}{c}{TED Emotion Dataset} \\ \cmidrule(r){2-7} \cmidrule(r){8-13}
\multirow{-2}{*}{\begin{tabular}[c]{@{}c@{}}Ablation\\ Settings\end{tabular}} & L2$\downarrow$ & MPJRE$\downarrow$ & FGD$\downarrow$ & BA$\uparrow$ & EA(\%)$\uparrow$ & Diversity$\uparrow$ & L2$\downarrow$ & MPJRE$\downarrow$ & FGD $\downarrow$ & BA$\uparrow$ & EA(\%)$\uparrow$ & Diversity$\uparrow$ \\ \midrule
Baseline & 1.99 & 3.52 & 1.16 & 0.90 & 59.43 & - & 1.36 & 3.47 & 1.29 & 0.90 & 59.59 & - \\
Baseline$\ddagger$ & 1.97 & 3.52 & 1.11 & 0.91 & 68.73 & - & 1.35 & 3.46 & 1.29 & 0.90 & 64.46 & - \\
+ EAD & 1.94 & 3.44 & 0.98 & 0.91 & 61.84 & - & 1.30 & 3.21 & 0.92 & 0.91 & 66.13 & - \\
+ $\mathcal{L}_{emo}$ & 1.88 & 3.32 & 0.83 & 0.91 & 72.35 & - & 1.25 & 2.95 & 0.79 & 0.91 & 76.62 & - \\
+ $\mathcal{L}_{beat}$ & 1.76 & 3.03 & 0.67 & 0.92 & 75.90 & - & 1.05 & 2.78 & 0.52 & 0.92 & 81.60 & - \\
+ Spatial & 1.73 & 2.97 & 0.61 & 0.92 & 78.89 & - & 0.99 & 2.60 & 0.40 & 0.92 & 82.91 & - \\
+ Temporal & 1.68 & 2.89 & 0.55 & 0.93 & 80.42 & - & 0.94 & 2.41 & 0.30 & 0.92 & 85.35 & - \\
\rowcolor[HTML]{ECF4FF} 
\textbf{+ $\mathcal{L}_{smooth}$} & \textbf{1.59} & \textbf{2.69} & \textbf{0.47} & \textbf{0.93} & \textbf{81.21} & \textbf{-} & \textbf{0.93} & \textbf{2.37} & \textbf{0.11} & \textbf{0.93} & \textbf{85.59} & \textbf{-} \\
\rowcolor[HTML]{ECF4FF} 
\textbf{+ Sampling} & \textbf{-} & \textbf{-} & \textbf{0.52} & \textbf{0.91} & \textbf{81.16} & \textbf{39.34$^{\pm 1.02}$} & \textbf{-} & \textbf{-} & \textbf{0.13} & \textbf{0.91} & \textbf{84.81} & \textbf{19.46$^{\pm 0.19}$} \\ \bottomrule
\end{tabular}%

\end{table*}

\begin{figure*}
\begin{center}
\includegraphics[width=0.95\linewidth]{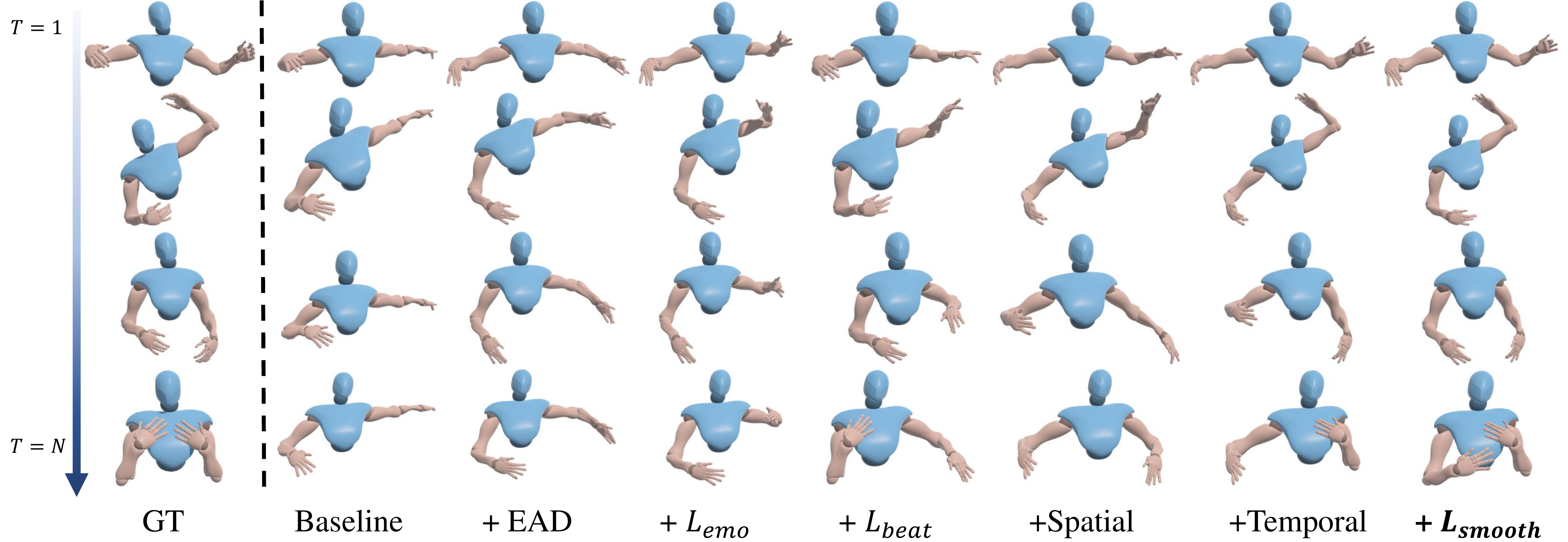}
\end{center}
\caption{ Visual comparisons of ablation study. We show the key frames of the generated gestures. From top to bottom, we show four key frames (an early, two middle, and a late one) of a pose sequence.
Best view on screen.}
\label{fig:figure5}
\end{figure*}

\subsubsection{Implementation Details.}
We set the whole length of the clip as $N=60$, and the length of initial poses as $M=10$ for both datasets. Similar to \cite{liu2022learning, ao2022rhythmic}, we employ ResNetSE-34~\cite{chung2020defence} as our audio encoder.  We adopt three stacking blocks and leverage the 2D-convolution-based header to map the dimension of audio features to be $N\times 512$, where $N$ is the temporal dimension. Then, we employ two MLP-based projectors to extract the emotion features and audio beat features. For initial poses, we propose an MLP-based pose encoder $E_{p}$ to obtain the embedded initial pose features. Then, we leverage a 1D convolution-based pose predictor to produce the future pose sequence.

For both BEAT and TED Emotion datasets, the frequency of speech audio is pre-processed to 16kHZ. Then, for more compact signal information preservation, the audio signals are converted to mel-spectrograms with the FFT window size is 1024 and hop length is 512. Finally, the audio signals are represented as the 2D time-frequency mel-spectrogram of size $ 128\times 124$.
Similar to \cite{ng2021body2hands, jiang2022avatarposer}, we convert the original 3D joints of both datasets into the 6D rotation representations~\cite{zhou2019continuity}. 
The 6D rotation representation has proved effective for training neural
networks due to its continuity. 

We set feature dimension $D = 512$, and $L=10$. Empirically, we set $\lambda_{r}=100$, $\lambda_{b}=0.05$, $\lambda_{e}=0.1$, $\lambda_{s}=0.5$, $\tau = 0.1$ in Eq. (\ref{Eq2}), and $\Gamma = 10$ in Eq. (\ref{Eq5}). 
Our model is implemented on the PyTorch platform with 2 NVIDIA RTX 2080Ti GPUs.  We adopt the Adam optimizer with an initial learning rate of 0.0002. The whole training takes 100 epochs with a batch size of 128. 

\subsubsection{Evaluation Metrics.}
To fully evaluate the superior performance of our proposed EmotionGesture framework, we employ the following metrics: 
\begin{itemize}
    \item \textbf{$L2$ Distance}: Distance between the generated co-speech gestures and target ones.
    \item \textbf{MPJRE}: Mean Per Joint Rotation Error $\left [ \circ  \right ] $ (MPJRE) \cite{jiang2022avatarposer, qi2023diverse} measures the absolute distance between the systhesized 3D representation joints and the pseudo ground truth.
    \item \textbf{FGD}: Fréchet Gesture Distance (FGD) \cite{yoon2020speech} evaluates the distribution distance between the ground truth and synthesized gesture. We pre-train an autoencoder based on BEAT and TED Emotion datasets to compute this metric.
    \item \textbf{BA}: Beat Alignment (BA) score \cite{liu2022beat} measures the rhythmic matching degree between the generated gestures and input audio.
    \item \textbf{EA}: Emotion Accuracy (EA) score measures whether the generated co-speech gestures represent the same emotion with audio. It is calculated by our pre-trained gesture-based emotion classifier.
    \item \textbf{Diversity}: Diversity score \cite{lee2019dancing, liu2022audio} indicates the difference among the generated gestures.   
\end{itemize}

\subsection{Quantitative Results}

\subsubsection{Comparisons with the state-of-the-art.}
We compare our EmotionGesture with previous state-of-the-art co-speech gesture generation approaches in recent years. We observe that the existing research could be roughly divided into two categories: 1) Directly generated the constant results from input audio signals, \ie, Seq2Seq~\cite{yoon2019robots}, S2G~\cite{ginosar2019learning}, JointEM\cite{ahuja2019language2pose}, and CAMN~\cite{liu2022beat}. 2) Generate the diverse co-speech gestures upon audio signals, \ie, Trimodal~\cite{yoon2020speech}, HA2G~\cite{liu2022learning}, DiffGes~\cite{zhu2023taming}, and TalkShow~\cite{yi2022generating}. Our framework can achieve optimal results in both experimental settings. For fair comparisons, all the models are implemented by source codes or pre-trained models released by authors. The experimental 
results of the counterparts are re-implemented under the same setting as ours.
The comparisons are divided into two parts.

First, we adopt the $L2$ Distance, MPJRE, FGD, BA, and EA for a well-rounded view of the constant generation comparisons. As reported in Table~\ref{table1}, our framework outperforms all the competitors with a large marginal gap. For instance, we surpass all methods on the metrics FGD, BA, and EA with both two datasets. 
Remarkably, on the BEAT dataset, our framework is even 56.9\% (\ie, $(1.09-0.47) / 1.09 \approx 56.9\%$) lower than the sub-optimal method on the FGD metric.
This indicates our synthesized co-speech gestures are much more realistic than other counterparts.
Meanwhile, the highest scores of BA and EA demonstrate our emotion-beat miming module facilitates the generated gestures to be frame-wise rhythmic while preserving emotion control.
As for the diversity result comparison, we exploit the FGD, BA, EA, and Diversity metrics to verify the superior performance of our framework. Our framework exceeds all the counterparts markedly from the perspective of diversity. Although the FGD, BA, and EA are slightly worse than the constant experimental setting due to the sampled emotion features, our framework still achieves optimal results. 
Meanwhile, we observe that the diversity scores of DiffGes~\cite{zhu2023taming} are much closer to ours due to being well adapted by the diffusion model~\cite{ho2020denoising}. However, the huge inference time makes this model demonstrate poor practical values in the real-time co-speech gesture applications.

\begin{table*}
\centering
\caption{Ablation study on the influence of transition chunk length $L$ in the spatial-temporal prompter. $\downarrow$ indicates the lower the better, and $\uparrow$ indicates the higher the better. "Duplicate" indicates repeating the initial poses to achieve the same temporal dimension as the target ones. Similarly, "Zero" means directly padding the temporal dimension of initial poses with zero elements.
Notice that in the padding strategy, we drop the spatial-temporal prompter in the experiments.}
\label{table5}
\setlength{\tabcolsep}{5 mm}
\footnotesize
\renewcommand{\arraystretch}{1.15} 
\begin{tabular}{llccccc}
\toprule
 &  & \multicolumn{5}{c}{BEAT Dataset} \\ \cmidrule(r){3-7} 
\multirow{-2}{*}{Variant Model} & \multirow{-2}{*}{Setting} & L2 $\downarrow$ & MPJRE $\downarrow$ & FGD $\downarrow$ & BA $\uparrow$ & EA(\%) $\uparrow$ \\ \midrule
 & Zero & 2.11 & 3.49 & 1.00 & 0.21 & 65.38 \\
\multirow{-2}{*}{Padding} & Duplicate & 1.96 & 3.39 & 0.78 & 0.88 & 73.78 \\ \midrule
 & $L=2$ & 1.69 & 2.91 & 0.57 & 0.91 & 76.10 \\
 & $L=4$ & 1.62 & 2.78 & 0.54 & 0.91 & 76.56 \\
 & $L=6$ & 1.61 & 2.76 & 0.52 & 0.92 & 78.13 \\
 & $L=8$ & 1.60 & 2.69 & 0.49 & 0.92 & 78.88 \\
\multirow{-5}{*}{Chunk Length} & \cellcolor[HTML]{ECF4FF}\textbf{$L=10$}& \cellcolor[HTML]{ECF4FF}\textbf{1.59} & \cellcolor[HTML]{ECF4FF}\textbf{2.69} & \cellcolor[HTML]{ECF4FF}\textbf{0.47} & \cellcolor[HTML]{ECF4FF}\textbf{0.93} & \cellcolor[HTML]{ECF4FF}\textbf{81.21} \\ \bottomrule
\end{tabular}%
\end{table*}

\begin{table*}[t]
\centering
\caption{The user study on naturalness, smoothness, and audio-gesture synchrony. The rating score range is 1-5, with 5 being the best. $\uparrow$ indicates the higher the better.}
\label{table3}
\renewcommand{\arraystretch}{1.1} 
\begin{tabular}{lccccccccc}
\toprule
Methods & Seq2Seq\cite{yoon2019robots} & S2G\cite{ginosar2019learning} & JointEM\cite{ahuja2019language2pose} & CAMN\cite{liu2022beat} & Trimodal\cite{yoon2020speech} & HA2G\cite{liu2022learning} & DiffGes\cite{zhu2023taming} & TalkShow\cite{yi2022generating} & \textbf{Ours} \\ \midrule
Naturalness $\uparrow$ & 2.92 & 2.33 & 2.83 & 3.38 & 3.98 & 3.87 & 4.43 & 4.30 & \cellcolor[HTML]{ECF4FF}\textbf{4.56} \\
Smoothness $\uparrow$ & 2.91 & 2.63 & 2.74 & 3.11 & 3.39 & 2.89 & 3.71 & 4.45 & \cellcolor[HTML]{ECF4FF}\textbf{4.75} \\
Synchrony $\uparrow$ & 3.07 & 2.64 & 2.63 & 3.30 & 3.02 & 3.55 & 3.80 & 4.72 & \cellcolor[HTML]{ECF4FF}\textbf{4.80} \\ \bottomrule
\end{tabular}%
\end{table*}

\subsubsection{Ablation Study.} We conduct the ablation study to demonstrate the effectiveness of each proposed component and loss function, as displayed in Table~\ref{table2}. Our baseline model is implemented by a simple transformer-based encoder-decoder backbone without emotion-beat mining in two branches. "+ EAD" in Table~\ref{table2} means we only adopt the two MLP-based projectors without other constraints. 
As shown in Table~\ref{table2}, all the combinations of our proposed modules and loss functions have positive impacts on the co-speech gesture generation. 
Specifically, even if we directly adopt the two MLP-based projectors without any other auxiliary losses (\ie, "+ EAD"), our framework reaches competitive results than utilizing the one-hot emotion vector (\ie, "Baseline$\ddagger$"). After we add the emotion classification loss $\mathcal{L}_{emo}$, the EA score is significantly improved on both datasets, 
Then, adopting the beat-alignment loss $\mathcal{L}_{beat}$ ideally improves the performance on $L2$ distance, MPJRE, FGD, and BA metrics. This supports our insight on "onsets start when people are speaking".

To verify the effectiveness of our proposed STP, we split it into two steps in the ablation. First, we just leverage the spatial-interpolation prompt learning strategy, our framework achieves the lower $L2$ distance, MPJRE, and higher EA. Then, we employ the temporal-reinforcement prompt learning to reach better performance, especially on the metrics of $L2$ distance, MPJRE. Both of these two components encourage the generated gestures to be more consistent with initial poses. Next, we adopt our simple yet effective motion-smooth loss $\mathcal{L}_{smooth}$. As noticed in Table~\ref{table2}, the FGD metric on both datasets decreased drastically. This indicates that $\mathcal{L}_{smooth}$ ensures the generated gestures realize realistic smooth gesture distribution. 

Additionally, to fully verify the effectiveness of the spatial-temporal prompter, we conduct ablation experiments on transition chunk length $L$. For fair comparisons, the experiments are implemented on the BEAT dataset without sampling during the inference phase, as reported in Table~\ref{table5}. The padding strategy achieves the much lower performance than our proposed spatial-temporal prompter, especially on the BA score. This suggests that directly padding initial poses as the pose prompt would lead to the generated gestures being low-fidelity and mismatched with the audio rhythmic beat. Although our prompt enhancement strategy is slightly influenced by the chunk length, it still significantly surpasses the direct padding upon the same length initial pose sequence. Even if our chunk length is 2, we still achieve better performance. 

\subsection{Qualitative Evaluation}

\subsubsection{User Study.}
Moreover, we conduct the user study based on the recruited 15 volunteers to better analyze the visual quality of the generated co-speech 3D gestures by various methods. Inspired by \cite{zhu2023taming, liu2022learning, liu2022audio}, our user study adopts \emph{naturalness, smoothness, and synchrony} as the main evaluation perceptions with the rating score range being 1-5 (the higher, the better). As displayed in Table~\ref{table3}, the average statistical results show that our framework gains the best performance in all three metrics. Especially in the \emph{naturalness} and \emph{synchrony}, our framework achieves significant advantages over all the other counterparts. This strongly proves the effectiveness of our proposed beat-alignment strategy and motion-smooth loss.

\subsubsection{Visualization.}

To fully verify the performance of our framework, we show the keyframes visualization of generated co-speech 3D gestures among our results against all the competitors in Figure~\ref{fig:figure4}. We observe that the Seq2Seq, S3G, and JointEM generate unnatural co-speech gestures compared with the ground truth. CAMN, Trimodal, and HA2G synthesize the natural gestures while their gestures are misaligned with the audio rhythm. 
In addition, although the DiffGes and TalkShow can generate better gestures than others, they stiffly match the audio signal by swinging the arms. 
Meanwhile, the gestures generated by these methods are less emotion-coherency with the audio. On the contrary, our framework reaches vivid and diverse co-speech gesture generation. As depicted in Figure~\ref{fig:figure4} and Figure~\ref{fig:figure1}, our framework enables the natural and diverse emotional gesture synthesis (\ie, neutral-to-happiness, the gestures swing at a wider angle). 

Moreover, to demonstrate the effectiveness of our proposed different components and loss functions, we visualize the keyframes of the generated co-speech gestures.
As illustrated in Figure~\ref{fig:figure5}, we can clearly observe that each component and loss function have a positive impact on the visualization of generated gestures. More demo videos are shown in our project page: \href{https://xingqunqi-lab.github.io/Emotion-Gesture-Web/}{\textit{https://xingqunqi-lab.github.io/Emotion-Gesture-Web/}}.



\section{Conclusion}

In this paper, we propose a novel framework \textbf{\emph{EmotionGesture}} to generate audio-driven vivid and diverse emotional co-speech 3D gestures. To achieve the audio-driven emotion control of the generated gestures, we fully take advantage of the audio-coherency transcripts for obtaining the emotional audio representation via the emotion-beat mining module. Then, we propose a spatial-temporal prompter to maintain the smoothness of the generated co-speech gestures. Moreover, we design a simple yet effective motion-smooth loss to smooth the jittering movement of the generated results. Extensive experiments on the BEAT and our newly collected TED Emotion datasets demonstrate the superior performance of our work with competitive ones. 

Our framework may produce some failure cases for some rarely-seen emotions (\eg, the disgust and contempt emotional gestures may not be easy to distinguish). We will explore handling the rarely-seen emotional gestures with a better generalization ability model in the future. Considering the broader impact, the generated co-speech gestures may be mis-leveraged in malicious avatar forgery. However, we believe our proposed technique would facilitate the research on multi-modality learning in a proper way of the real applications.

\bibliographystyle{IEEEtran}
\bibliography{Template}

\vfill

\end{document}